\documentclass[10pt,twocolumn,letterpaper]{article}

 \usepackage[pagenumbers]{cvpr} %

\usepackage[dvipsnames]{xcolor}

\usepackage{multirow}

\definecolor{cvprblue}{rgb}{0.21,0.49,0.74}
\usepackage[pagebackref,breaklinks,colorlinks,citecolor=cvprblue]{hyperref}
\usepackage{xcolor,colortbl}
\definecolor{Gray}{gray}{0.85}
\newcolumntype{a}{>{\columncolor{Gray}}c}

\title{OccFlowNet: Towards Self-supervised Occupancy Estimation via Differentiable Rendering and Occupancy Flow}

\author{Simon Boeder \quad Fabian Gigengack \quad Benjamin Risse \\
Robert Bosch GmbH, University of Münster \\
\texttt{\{simon.boeder, fabian.gigengack\}@de.bosch.com; b.risse@uni-muenster.de;} }

\begin{document}
\maketitle

\begin{abstract}
Semantic occupancy has recently gained significant traction as a prominent 3D scene representation.
However, most existing methods rely on large and costly datasets with fine-grained 3D voxel labels for training, which limits their practicality and scalability, increasing the need for self-monitored learning in this domain.
In this work we present a novel approach to occupancy estimation inspired by neural radiance field (NeRF) using only 2D labels, which are considerably easier to acquire.
In particular, we employ differentiable volumetric rendering to predict depth and semantic maps and train a 3D network based on 2D supervision only.
To enhance geometric accuracy and increase the supervisory signal, we introduce temporal rendering of adjacent time steps.
Additionally, we introduce occupancy flow as a mechanism to handle dynamic objects in the scene and ensure their temporal consistency.
Through extensive experimentation we demonstrate that 2D supervision only is sufficient to achieve state-of-the-art performance compared to methods using 3D labels, while outperforming concurrent 2D approaches.
When combining 2D supervision with 3D labels, temporal rendering and occupancy flow we outperform all previous occupancy estimation models significantly.
We conclude that the proposed rendering supervision and occupancy flow advances occupancy estimation and further bridges the gap towards self-supervised learning in this domain.
\end{abstract}
    
\section{Introduction}
\label{sec:intro}
Autonomous driving poses a significant challenge in the field of computer vision.
A crucial aspect of developing self-driving systems is the perception of the environment and its objects~\cite{survey1,survey2,survey3}.
Most frequently, supervised object detection methods identify known objects such as cars or pedestrians based on visual 2D and 3D data~\cite{survey1,survey2,survey3,huang2021bevdet,li2022bevformer,hu2021fiery,wang2023exploring, chen2016monocular, yolo2023glenn}, hence requiring curated datasets with predefined object classes.
For autonomous driving to be considered safe, it is however crucial to identify any object that might collide with the vehicle in all directions, ensuring $360^\circ$ coverage. 
This includes not only objects encountered during training but also any unforeseen objects and therefore necessitates solutions for \textit{generic object detection}.
Detecting objects with arbitrary appearances and shapes in unconstrained scenes requires generic representations such as 3D occupancy in a voxel-grid, which has gained considerable popularity recently~\cite{roldao20223d,cao2022monoscene, chen2023end,huang2023tri,tong2023scene,zhang2023occformer,li2023fb, tian2023occ3d,wang2023openoccupancy,philion2020lift,peng2023learning}.
As a result, numerous occupancy prediction methods have been introduced, necessitating large and extensively annotated datasets for effective training~\cite{tong2023scene,wang2023openoccupancy, tian2023occ3d}. 
This has led to a growing need for more adaptable methods that utilize easily obtainable data and require less effort in annotation.
For example, recent methods have attempted to avoid 3D labels \cite{pan2023renderocc,huang2023selfocc}, but could not meet the performance of 3D-based methods.

In this paper, we propose a novel method called OccFlowNet, which is the first approach to achieve state-of-the-art 3D occupancy estimation performance by using 2D supervision only.
We utilize differentiable volumetric rendering inspired by NeRF (Neural Radiance Fields)~\cite{mildenhall2021nerf} to train a 3D occupancy network using only images and semantic LiDAR points, without the need for 3D voxel labels.
In particular, we extend the concepts of bird's-eye-view (BEV) perception~\cite{huang2021bevdet, li2023bevstereo} to predict 3D semantic voxel fields, and train them in 2D space by rendering depth and semantic maps from the predictions and supervising it using 2D labels obtained from LiDAR scans.
If available, 3D labels can still be integrated into OccFlowNet yielding even higher performance values.
To provide more overlapping supervision, temporally adjacent frames are also rendered, since the intersection between rays of adjacent cameras is typically low~\cite{caesar2020nuscenes}.
Additionally, we introduce occupancy flow as a mechanism to handle dynamic objects during temporal rendering when 3D bounding box annotations are available.

To demonstrate the potency of our proposed approach, we conduct extensive experiments on the widely used Occ3D-nuScenes dataset \cite{tian2023occ3d} and compare to recent approaches using both 2D and/or 3D supervision.
The results indicate that our approach, using only 2D supervision, can yield comparable and often better results than other state-of-the-art 3D occupancy prediction models, and outperforms all previous attempts for 2D supervised occupancy estimation.
Moreover, our approach sets the new state-of-the-art performance on the Occ3D-nuScenes~\cite{tian2023occ3d} benchmark by combining 2D and 3D supervision, outperforming all previous methods.
A detailed ablation study further investigates the reasons for the success of our method, providing more insight into the proposed components.
In summary, our contributions are:
\begin{enumerate}
    \item Differentiable volume rendering as an alternative training paradigm, eliminating the need for 3D labels.
    \item Use of temporal rendering and occupancy flow to handle dynamic objects, lifting the performance of 2D supervision to the level of 3D-based methods.
    \item Demonstrating that combined 2D and 3D supervision achieves the up to now best performance on the widely used Occ3D-nuScenes benchmark.
\end{enumerate}
\section{Related Work}
\label{sec: related work}
In the context of our proposed algorithm, three related areas of research are of particular importance, namely Vision-based 3D Object Detection, Occupancy Estimation and Differentiable Rendering and Neural Radiance Fields. 

\subsection{Vision-based 3D Object Detection}
The task of vision-based 3D object detection is a prominent and well studied subject in the fields of autonomous vehicles, robotics, and augmented reality, primarily due to the widespread availability of image data~\cite{survey2, survey3, caesar2020nuscenes}.
The objective of this task is to identify and locate objects in 3D space using camera images.
State-of-the-art methods encompass a vast range of approaches, including monocular~\cite{chen2016monocular, peng2023learning,wang2021fcos3d,Simonelli_2019_ICCV}, and multi-view detection~\cite{huang2021bevdet, li2022bevformer, hu2021fiery, li2023bevstereo, huang2022bevdet4d, wang2021fcos3d, park2022time, liu2022petr}.
To predict objects in 3D space, most models employ a transformation of 2D image features into a 3D representation.
A common approach is to transform features from multiple images into a shared Birds-Eye-View (BEV) grid.
Methods such as LSS~\cite{philion2020lift} and derived approaches~\cite{huang2021bevdet, li2023bevstereo, huang2022bevdet4d} predict depth distributions to project image features onto the BEV grid and estimate 3D bounding boxes based on these features.
In a different approach introduced by BEVFormer~\cite{li2022bevformer}, a set of learnable BEV grid queries are projected onto image features to calculate interactions.
Methods like DETR3D~\cite{wang2022detr3d} and StreamPETR~\cite{wang2023exploring} follow an object-centric approach for detection and define queries on an object level.

Nevertheless, classical 3D object detection methods can only detect object classes present in the training data which has led to the emergence of the occupancy estimation paradigm.
Building upon the practicality of BEV-based methods, our proposed approach utilizes the prominent BEVStereo method~\cite{li2023bevstereo} to transform image features for the 3D occupancy estimation task.

\subsection{Occupancy Estimation}
The SemanticKITTI challenge~\cite{behley2019semantickitti} introduced the first dataset for the occupancy estimation using LiDAR data and a stereo camera setting.
More recently, the Occ3D-nuScenes~\cite{tian2023occ3d} and OpenOccupancy datasets~\cite{wang2023openoccupancy} provided a fine-grained 3D occupancy ground truth for the nuScenes dataset~\cite{caesar2020nuscenes}. 
In contrast to object detection, where specific bounding boxes have to be predicted, the objective is to estimate the 3D geometry and semantics of the entire scene within a predefined voxel grid.
Initial works on the SemanticKITTI benchmark use LiDAR scans, monocular or stereo images as an input to estimate occupancy~\cite{cao2022monoscene,li2023voxformer,huang2023tri,miao2023occdepth}.
More recent approaches on the Occ3D-nuScenes benchmark utilize the multi-view camera setup to lift multiple images into a shared 3D space~\cite{wang2023panoocc,jiang2023symphonies,gan2023simple,li2023voxformer,huang2023tri,wei2023surroundocc}. These methods typically adopt architectures based on previous 3D object detection techniques~\cite{huang2021bevdet,li2022bevformer} and incorporate additional heads to predict occupancy~\cite{li2023voxformer,huang2023tri,zhang2023occformer,jiang2023symphonies,li2023fb,huang2022bevdet4d}.

However, due to the inherent complexity of dense 3D occupancy estimation, most existing methods rely on costly voxel-based 3D ground-truth labels for training.
In this work, we train a model solely using LiDAR scans as ground truth, eliminating the need for voxel labels.
Lately, there have been attempts to train occupancy networks without the use of 3D labels, in a way that closely resembles the method we propose.
The concurrent work RenderOcc~\cite{pan2023renderocc} also use differentiable rendering to supervise the occupancy in 2D space (at the time of submission RenderOcc is available as a pre-print only). 
Due to some structural similarities we provide a direct comparison in \cref{sec:diff_renderocc}.

\subsection{Differentiable Rendering and Neural Radiance Fields}
Training depth and detection models on vision-only input, known as self\-/supervised learning, has a rich history of research~\cite{guizilini2022full,zhou2017unsupervised,Godard_2019_ICCV},  leveraging reprojection errors and cost-volume regularization in temporal and multi-view scenarios.
The introduction of Neural Radiance Fields (NeRF)~\cite{mildenhall2021nerf} has pushed this paradigm further, enabling 3D scene reconstructions from RGB images.
This is achieved by employing differential volumetric rendering to train a scene-specific Multi-Layer Perceptron (MLP) that predicts density and color values for 3D locations and viewing directions.
While traditional NeRFs and their variants~\cite{barron2021mip, barron2022mip, mildenhall2021nerf, martin2021nerf, deng2022depth, muller2022instant} are typically fitted to a specific scene and require numerous posed images or extensive training times, subsequent works incorporate different priors like depth and image features, or additional data like LiDAR, to train generalizable NeRF models~\cite{xu2022point,chang2022rc,chen2021mvsnerf,yu2021pixelnerf}.

In this work, we propose to leverage the concept of differential volume rendering to train a robust and generalizable 3D perception model.
At the time of writing, other works are emerging that follow a similar direction.
RenderOcc~\cite{pan2023renderocc} also use volume rendering of depth and semantic maps with LiDAR ground truth to train a occupancy model, while SelfOcc~\cite{huang2023selfocc} and OccNeRF~\cite{zhang2023occnerf} use rendering to reconstruct RGB images to train a network in a fully self-supervised manner.
However, none of these methods explicitly address moving objects, despite the propagation of temporal information is key for each of these methods, which inevitably introduces false supervisory signals.
\section{Methodology}
\label{sec:methodology}
\subsection{Problem Definition}\label{sec:problem_definition}
Given a set of $N$ multi-view images $I = \{I^1, I^2, ..., I^N\}$ of the current time step, the aim of occupancy prediction is to correctly estimate the semantic voxel volume $V = \{c_0, c_1, ... c_{C'}\}^{X \times Y \times Z}$ on a predefined grid, where $C'$ is the number of semantic classes including a \textit{free} class that identifies empty space.
For estimating the volume, a model $\mathbb{M}$ is used, such that

\begin{align}
    \hat{V} = \mathbb{M}(I) \text{, } \quad I = \{I^1, I^2, ..., I^N\}
\end{align}

\subsection{Model architecture}\label{sec:model_arch}
\begin{figure*}[t]
    \centering
    \includegraphics[page=1, trim=0cm 6.9cm 0.37cm 0cm, clip, width=1\textwidth]{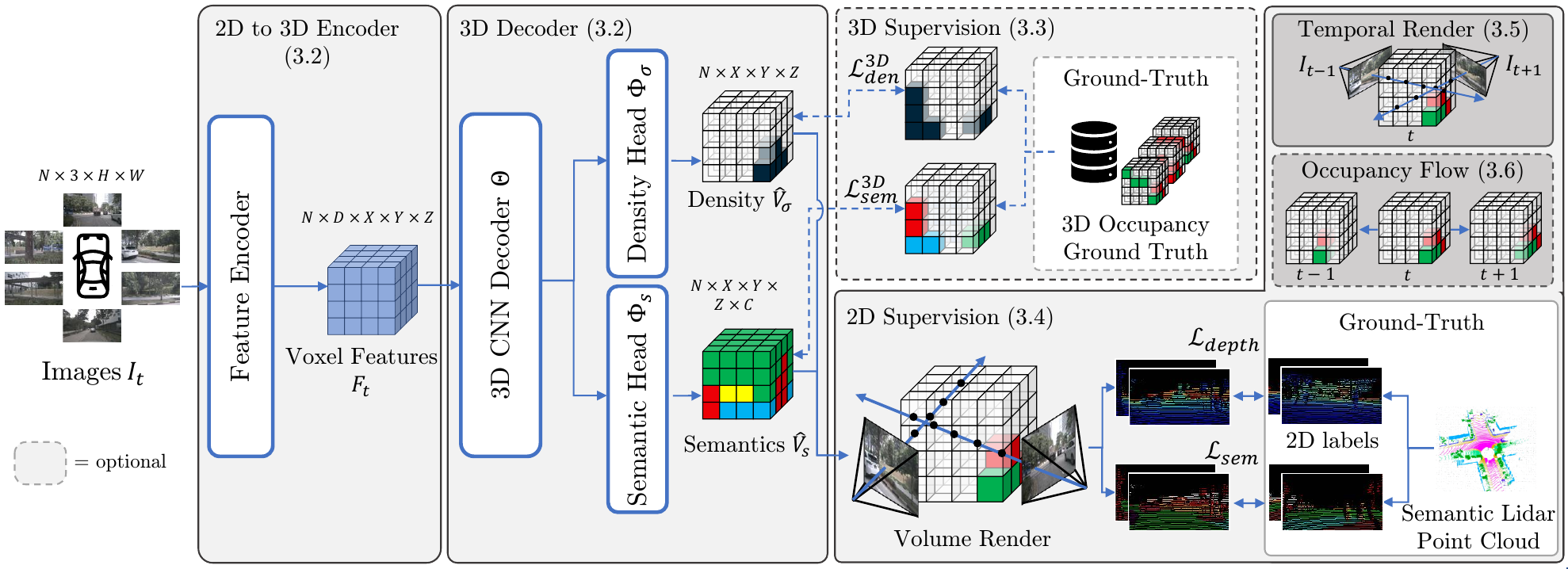}
    \caption{
        \textbf{Complete architecture of the proposed model.}
        Multi-view images are used as an input to generate density and semantic predictions in a predefined voxel grid.
        The model is trained by rendering 2D depth and semantic maps from its predictions and by computing a loss on depth and semantic ground truth obtained from LiDAR scans (\ref{sec:selfsupervised}).
        Optionally, the model can be trained using 3D voxel labels (\ref{sec:supervised}).
        \textit{N, H, W} denote the number of images, the image height and the image width and \textit{X, Y, Z} represent the voxel grid dimensions.
        \textit{D, C} are the latent dimension size and number of classes, respectively.
    }
    \label{fig:architecture}
\end{figure*}

The architecture of the proposed model $\mathbb{M}$, which transforms input images into semantic occupancy predictions, is depicted in \cref{fig:architecture}.
We use the BEVStereo~\cite{li2023bevstereo} model to transform image features into 3D voxel features as it has been shown to be able to extract very effective 3D features.
However, any 2D-to-3D encoder like~\cite{li2022bevformer, philion2020lift, huang2021bevdet, huang2023tri} could be used instead.
After creating the voxel features, the 3D decoder will estimate density probabilities and semantic logits per voxel using a combination of 3D CNNs and MLPs.
Finally, in order to supervise the model, we can either use 3D voxel ground truth if available (\cref{sec:supervised}) or conduct volume rendering to train with 2D labels (\cref{sec:selfsupervised}).

\paragraph{2D-to-3D Feature Encoder}\label{sec:bevstereo}
Our proposed method is based on BEVStereo~\cite{li2023bevstereo}, which is one of the current state-of-the-art multi-view 3D object detection models.
It takes as input the set of images at the current time step as well as some previous frames for temporal information and outputs a set of 3D features that are pooled to the BEV plane, from which a task-specific head will predict 3D object boxes.

This is done by using a pre-trained backbone architecture $B$ to extract image features. 
Subsequently, these features are projected into 3D space (camera frustums) using a depth estimation similar to the well-known Lift-Splat-Shoot architecture~\cite{philion2020lift}, referred to here as $LSS$.
Instead of pooling these features on the BEV plane, voxel pooling $P$ is used to obtain features on a 3D voxel grid.
Moreover, the 3D object detection head is removed, as we will deploy a custom 3D decoder that predicts 3D occupancy.
The extraction of voxel features $F_t$ via BEVStereo can thus be summarised as:

\begin{align}
F_t = P(LSS(B(I))) \; .
\end{align}

\paragraph{3D Decoder}\label{sec:decoder}
The 3D voxel decoder takes the voxel features $F_t$ from the feature encoder network as an input and estimates the density and the semantic label at each voxel.
First, the voxel features are further refined and locally aggregated by employing a 3D decoder $\Theta$ of multiple blocks, where each block consists of a set of 3D convolutional layers, batch norm layers, ReLU activations and skip connections.
The authors of BEVDet~\cite{huang2021bevdet} and BEVStereo~\cite{li2023bevstereo} use the same decoder architecture for their occupancy network variants.
Afterwards, two separate MLP networks $\Phi_\sigma$ and $\Phi_s$ estimate the occupancy probability $\hat{V}_\sigma$, which can also be interpreted as the volume density and the semantic logits $\hat{V}_s$ for each voxel for the set of classes $C$.
Here $C$ does not contain the free space class $c_{free}$ since the occupancy probability is estimated using $\Phi_\sigma$.
To transform the density logits of $\Phi_\sigma$ to probabilities, we use the sigmoid function, denoted as $\psi$. 

\begin{align}
  &\hat{V}_s = \Phi_s( \Theta( F_t)) \in \mathbb{R}^{X \times Y \times Z \times C} \; \\
  &\hat{V}_\sigma = \psi ( \Phi_\sigma(\Theta(F_t))) \in [0, 1]^{X \times Y \times Z}
  \label{eq:decoder}
\end{align}

At inference, the semantic logits and the estimated density probabilities can be combined to create the final semantic occupancy prediction $\hat{V}$ given a confidence threshold $\tau$ as follows:
\begin{align}
  \hat{V}(x,y,z) = 
  \left\{\begin{array}{lr}
        argmax(\hat{V}_s(x,y,z)), & \text{ for } \hat{V}_\sigma(x,y,z) \geq \tau \\
        c_{free}, & \text{ for } \hat{V}_\sigma(x,y,z) < \tau \\
        \end{array}
  \right. 
  \label{eq:decoder_combined}
\end{align}

\subsection{3D Supervision}\label{sec:supervised}
In contrast to other works~\cite{huang2023tri, tian2023occ3d,pan2023renderocc,zhang2023occformer} our model can be trained without 3D voxel labels (see \cref{sec:selfsupervised}).
Nevertheless, these labels can also be integrated in our architecture as shown in \cref{fig:architecture}.
This is done by defining two losses $\mathcal{L}_{den}^{3D}$ and $\mathcal{L}_{sem}^{3D}$ to supervise the outputs of $\Phi_\sigma$ and $\Phi_s$ directly.
Given ground truth voxel labels $V = \{c_0, c_1, ... c_{C'}\}^{X \times Y \times Z}$, we can define the occupancy probabilities and the semantic labels as follows:

\begin{align}
&V_\sigma (x,y,z) =   
    \left\{\begin{array}{lr}
        0,  & \text{ for } V(x,y,z) = c_{free} \\
        1, & \text{ otherwise}   \\
        \end{array}
    \right. \; , \\
&V_s (x,y,z) = 
    \left\{\begin{array}{lr}
        V(x,y,z),  & \text{ for } V(x,y,z) \neq c_{free} \\
        ignore, & \text{ otherwise}   \\
        \end{array}
    \right. \; .
\end{align}

The occupancy probability labels are therefore defined for the whole volume $V$, while semantic labels are only available for non-empty voxels (because $\Phi_s$ does not predict the free class).
We use the binary cross-entropy loss for $\mathcal{L}_{den}^{3D}$ and the cross-entropy loss for multi-class classification $\mathcal{L}_{sem}^{3D}$.
The final 3D supervised loss is the sum of both losses.

\subsection{Rendering Supervision in 2D}\label{sec:selfsupervised}
To enable training of $\mathbb{M}$ even when 3D voxel-based labels are not available, we propose to use differentiable volume rendering with 2D labels as a training mechanism.
This method became very popular recently with the rise of Neural Radiance Fields (NeRF)~\cite{mildenhall2021nerf}.
However, differentiable rendering can not only be used to fit a radiance field to a specific scene.
In this work, we render the estimated volume $\hat{V}$ in a differentiable manner to obtain estimated depth and semantic maps, which can be used to compute a loss based on 2D ground truth.

\paragraph{Ray Generation}\label{sec:ray_gen}
Given the extrinsic and intrinsic parameters $E_i$ and $K_i$ of the camera $i$, we can create a set of rays $\mathcal{R}$, where each ray $r$ is going from a camera origin through a pixel of the input image.
For a pixel location $(u, v)_i$, the ray origin $o_i(u, v)$ and ray direction $d_i(u, v)$ are created via the following equation:
\begin{align}
    o_i(u, v) = E_i, \quad
    d_i(u, v) = E_i K_1^{-1} [u, v, 1] \; .
\end{align}
During training, we will only create rays for $(u, v)$ locations that have ground truth depth and semantic information available, as we will explain later.
Given a ray $r$, we can sample a set of points $r(t) = o + td$ at different distances $t$ along the ray and use volume rendering as in~\cite{mildenhall2021nerf} to compute the depth $\hat{\mathcal{D}}(r)$ and semantic label $\hat{\mathcal{S}}(r)$.
We use the coarse-to-fine sampling approach described in~\cite{barron2021mip} to determine the sample distances $t$.

\paragraph{Volume Sampling}
For every sample $r(t)$ along the ray, we will fetch the occupancy probability $\sigma(r(t))$ and the semantic logits $s(r(t))$ by trilinear interpolating the estimated volumes $\hat{V}_\sigma$ and $\hat{V}_s$ at the 3D position of sample $r(t)$.

\paragraph{Volume Rendering}
A rendering weight $w(r(t))$ for each sample can be computed by accumulating the interpolated density along the ray:
\begin{align}
    &w(r(t)) = T\left(r(t)\right) \left(1 - \exp(- \sigma(r(t)) \delta_t) \right) \textrm{, with} \\
    &T(r(t)) = \exp\left(-\sum_{j=1}^{t-1} \sigma(r(j)) \delta_j \right)
\end{align}
where $T(r(t))$ represents the accumulated transmittance along the ray until $t$, and $\delta_t = r(t+1) - r(t)$ is the distance between two samples.
Finally, we can compute the rendered depth and semantics by summing up the respective values multiplied with the rendering weights along the ray.
\begin{align}
    &\hat{\mathcal{D}}(r) = \sum_{t=1}^N w(r(t)) t , \\
    &\hat{\mathcal{S}}(r) = \sum_{t=1}^N w(r(t)) s(r(t)) \; .
\end{align}

\paragraph{Losses}
Given the rendered depth and semantic logits, losses can be computed given some ground truth depth and semantic labels $\mathcal{D}(r)$ and $\mathcal{S}(r)$. 
We use the mean squared error as the depth loss $\mathcal{L}_{depth}$, the cross-entropy as the semantic loss $\mathcal{L}_{sem}$ and sum up these two losses to create the final loss.
In order to create ground truth depth and semantic labels, we project the annotated LiDAR scan from the current frame onto the image plane of the input cameras to obtain a set of $(u, v)$ locations and corresponding depth and semantic labels.
Note that the whole rendering supervision is only used during training. 
At inference, only the model $\mathbb{M}$ is used to predict the semantic occupancy volume $\hat{V}$.

\subsection{Temporal Rendering}\label{sec:temp_render}
Since adjacent cameras on autonomous vehicles usually share a relatively small frustum overlap, rays from different cameras will rarely intersect, aggravating consistent depth estimations.
To increase the total number of times each voxel is seen and therefore increase overall supervision, additional rays from temporally adjacent frames are generated (referred to as \textit{temporal frames}).
During training, for each frame, a number of past and future frames defined by horizon $O$ are selected and additionally loaded to create a set of frames $D_{temp}$ as 
\begin{align}
    &D_{temp} = \{ I_t \text{ }|\text{ } t \in T\} , \\
    &T = \{ t \text{ } | \text{ } t \in \mathbb{Z}, -O \leq t \leq O \} ,
\end{align}
where $T$ is the set of temporal indices.
For each frame $I_t$, the respective LiDAR point cloud $P_t$ is loaded and projected onto the images to generate the depth and semantic ground truth.
Then, training rays are generated from all temporal frames in the same way as explained in \cref{sec:selfsupervised}, and are transformed to the current frame via the known camera and ego poses.
Note that future frames are only used during training and not during inference.

Automated driving data usually contains dynamic objects.
If one naively generates rays from different time steps, objects in the scene might have moved, leading to wrong rays generated at temporal frames. 
Moreover, dynamic objects might reveal other objects that are occluded in the current frame, hence leading to additional wrong LiDAR points in the current frame.
To address these errors, we propose two filtering techniques to ensure that temporal frames are rendered correctly even in the presence of dynamic objects and a loss weighting to counteract the consequential class imbalance.

\paragraph{Dynamic Ray Filter}\label{sec:ray_filter}
In each temporal frame ($t \neq 0$), all LiDAR points belonging to dynamic classes are removed, so that no rays are generated for dynamic objects in temporal frames.
As a result, only the static world is rendered in temporal frames, which is consistent with the current frame.
The accumulated dataset of points for which rays are generated is thus defined as:
\begin{equation}
  D_P = \{ P_t^c | (t=0) \textrm{ or } (t \neq 0  \text{ and } c \in C_{static} ) \} \, ,
  \label{eq:ray_gen}
\end{equation}
where $P_t^c$ is a LiDAR point with label $c$ at time step $t$, and $C_{static}$ is the set of static class labels.

\paragraph{Disocclusions}\label{sec:disocclusions}
Since moving objects can cover or reveal static world points over time, the removal of dynamic classes will inevitably result in a small fraction of wrong rays for otherwise static points.
These so-called disoccluded points have a static class label as the dynamic object has moved, and therefore were not removed in the previous step.
However, the label is still wrong as we should predict the dynamic object for this ray.
To resolve this issue, we set the predicted density of all dynamic objects to $0$ in temporal frames, hence removing them from our predictions. 
This way, the correct background class can be predicted for each revealed point.

\paragraph{Loss Weighting}
Dynamic objects like cars or pedestrians are in general less frequent in driving scenes than static classes, but are usually of high importance for autonomous driving applications.
The dynamic label scarceness is amplified by filtering out dynamic objects at temporal frames.
To counteract this imbalance, we scale the loss of each ray by a weight that is computed based on the occurrence of its class, precomputed over the whole dataset.
The weight $\omega$ for a ray $r$ with class $c$ is computed as
\begin{equation}
  \omega(c) = \log \left( \frac{ \sum_{k=0}^C N_k}{N_c} \right)
  \label{eq:loss_weight}
\end{equation}
where $N_i$ is the number of points with label $i$ in the whole dataset.

\subsection{Occupancy Flow}\label{sec:occ_flow}
Training models with rendering supervision and temporal frames already yields accurate estimations of semantic occupancy (cf. \cref{sec:experiments}).
However, there is still a performance gap between static and dynamic classes due to the large label imbalance imposed by the dynamic object filtering.
To enable learning of the dynamic classes even at temporal frames, we introduce occupancy flow.
We precompute the scene flow of each dynamic voxel in the dataset to each target time step in the horizon $O$ based on ground truth 3D bounding boxes. 
With available voxel flow, we can move the estimated occupancy of dynamic objects to the correct positions in the temporal frames, and can render these correctly without having to filter them out, strengthening the supervision of dynamic objects. 

During training, we can load all dynamic voxel indices, their classes and their transformations to the temporal time steps.
The estimated occupancy is then copied $|T|$ times to create one estimation for each temporal time step. 
Using the indices and the class of the boxes, we determine which estimated voxels have to be moved. 
We only select voxels that were predicted correctly and are within the box, so that we do not accidentally move falsely estimated voxels.
Afterwards, the flow transformations are applied to the positions of all selected voxels, moving them in the respective copy of the occupancy estimation to the target position.
To realign the transformed points with the grid, inverse distance-based weighted interpolation is used.
Formally, given the position $x$ of a point after the flow transformation is applied, its nearest eight voxel centers $v_i$ on the predefined grid are selected. 
Then, for each adjacent voxel center, an interpolation weight is computed inversely proportional to the distance to $x$ as follows:
\begin{equation}
  w_i(x) = \frac{1}{|x - x_i|}
  \label{eq:idw}
\end{equation}
Using these weights, the semantic logits $s$ and density probability $\sigma$ of all adjacent voxel centers are updated to distribute the moved voxels values back to the grid:

\begin{equation}
  s(v_i) = \frac{w_i(x) }{\sum_{p=1}^P w_p(x)} * s(v) + \left(1 - \frac{ w_i(x)}{\sum_{p=1}^P w_p(x)} \right)  * s(v_i) \;
  \label{eq:update_interpol}
\end{equation}
where $p$ indexes all selected adjacent voxels centers and the same update formula is repeated for all $\sigma(v_i)$.

Afterwards, instead of single static fields, we now have $|T|$ semantic and density fields $\bar{V}_s \in \mathbb{R}^{|T| \times X \times Y \times Z \times C}$ and $\bar{V}_\sigma \in [0, 1]^{|T| \times X \times Y \times Z}$ that resemble the state of the scene in each temporal time step, with dynamic objects moved according to the scene flow. 
Volume rendering can be executed in the same way as explained above, except the rays are cast through the occupancy field of the respective time step. 
\section{Experiments}
\label{sec:experiments}

\subsection{Dataset}
We evaluate our OccFlowNet on the Occ3D-nuScenes benchmark~\cite{tian2023occ3d}, which provides semantic 3D voxel-based ground truth for the widely used nuScenes dataset~\cite{caesar2020nuscenes}. 
The voxel grid around the vehicle is defined in a range of $-40m$ to $40m$ on the X and Y-axis, and from $-1m$ to $5.4m$ on the Z-axis, given in the ego coordinate frame.
The voxel size is $0.4m$, resulting in an occupancy grid of size $200 \times 200 \times 16$.
Each voxel is assigned one of the 17 class labels from the nuScenes Lidarseg \cite{fong2021panoptic} dataset or the free class.
The performance on this benchmark is measured with the Intersection over Union (IoU) score of the 17 semantic categories, where mIoU is the mean over all classes.
Additionally, for the occupancy flow, we use the annotated 3D bounding boxes of the nuScenes dataset in order to precompute the voxel flow ground-truth.

\subsection{Implementation Details}
For all experiments, we use the SwinTransformer~\cite{liu2021swin} as the image backbone, with an image resolution of $512 \times 1408$.
The density decoder $\Phi_\sigma$ and semantic decoder $\Phi_s$ each consist of three hidden layers with dimensionality of 256.
We use the same optimizer settings as in BEVStereo4D~\cite{li2023bevstereo} and train for 12 epochs with a batch size of 4, on 4 NVIDIA A100 GPUs.
We employ a time horizon $O=3$ for temporal rendering, and sample $32,768$ rays randomly for each frame.
For each ray, we sample $50$ proposal samples and $100$ fine samples for volume rendering and use the \textit{nerfacc}~\cite{li2023nerfacc} package for sampling and rendering.
When training with rendering supervision (\cref{sec:selfsupervised}), the 2D labels are generated by projecting LiDAR points on the camera images.
For training with 3D labels (\cref{sec:supervised}), we use the ground truth data of Occ3D-nuScenes, and do \textbf{not} use the camera mask, as recommended by the authors.

\subsection{3D Occupancy Prediction Results}
We evaluate our proposed method against common state-of-the-art approaches for occupancy prediction on the Occ3D-nuScenes dataset~\cite{tian2023occ3d} and the results are summarized in \cref{table:main}.
We provide three configurations of our model.
The \textit{Base} model where only rendering supervision is used, with temporal rendering and dynamic filtering.
The \textit{Flow} model, when adding the occupancy flow component.
Lastly, we also provide results for our model when also using 3D voxel labels.

\begin{table*}
\begin{center}
\caption{
\textbf{Occupancy estimation performance on the Occ3D-nuScenes validation set.}
Performance is measured in \%IoU, best performing per column in \textbf{bold}, second best in \textit{italics}.
Our method outperforms most 3D methods even with 2D training only.
Using 3D labels, we surpass all competitors.
Dynamic object categories are highlighted in \colorbox{Gray}{gray}.}
\label{table:main}

\resizebox{\textwidth}{!}{%
\addtolength{\tabcolsep}{2pt}
    \begin{tabular}{lc|c|ccaaaaaacaacccccc}
    \hline
    \noalign{\smallskip}
    Method & \rotatebox{90}{Mode} & mIoU & \rotatebox{90}{others} & \rotatebox{90}{barrier} & \rotatebox{90}{bicycle} & \rotatebox{90}{bus} & \rotatebox{90}{car} & \rotatebox{90}{cons. vehicle} & \rotatebox{90}{motorcycle} & \rotatebox{90}{pedestrian} & \rotatebox{90}{traffic cone} & \rotatebox{90}{trailer} & \rotatebox{90}{truck} & \rotatebox{90}{driv. surf.} & \rotatebox{90}{other flat} & \rotatebox{90}{sidewalk} & \rotatebox{90}{terrain} & \rotatebox{90}{manmade} & \rotatebox{90}{vegetation}\\
    \noalign{\smallskip}
    \hline
    \noalign{\smallskip}
    OccFormer \cite{zhang2023occformer} & 3D & 21.93 & 5.9 & 30.3 & 12.3 & 34.4 & 39.2 & 14.4 & 16.4 & 17.2 &  9.3 & 13.9 & 26.4 & 51.0  & 31.0  & 34.7 & 22.7 &  6.8 &  7.0 \\
    BEVFormer \cite{li2022bevformer} & 3D & 23.67 & 5.0  & 38.8 & 10.0  & 34.4 & 41.1 & 13.2 & 16.5 & 18.2 & 17.8 & 18.7 & 27.7 & 49.0  & 27.7 & 29.1 & 25.4 & 15.4 & 14.5 \\
    BEVStereo \cite{li2023bevstereo} & 3D & 24.51 & 5.7 & 38.4 &  7.9 & 38.7 & 41.2 & 17.6 & 17.3 & 14.7 & 10.3 & 16.8 & 29.6 & 54.1 & 28.9 & 32.7 & 26.5 & 18.7 & 17.5 \\
    RenderOcc \cite{pan2023renderocc} & 2D+3D  & 26.11 & 4.8 & 31.7 & 10.7 & 27.7 & 26.4 & 13.9 & 18.2 & 17.7 & 17.8 & 21.2 & 23.2 & 63.2 & 36.4 & \textbf{46.2} & \textbf{44.3} & 19.6 & 20.7\\
    FB-Occ \cite{li2023fb} & 3D & 27.09 & 0.0  & \textbf{40.9} & 21.2 & 39.2 & 40.8 & \textit{20.6} & 23.8 & \textbf{23.6} & \textit{25.0}  & 16.6 & 26.4 & 59.4 & 27.6 & 31.4 & 29.0  & 16.7 & 18.4 \\
    TPVFormer \cite{huang2023tri} & 3D & 27.83 & 7.2 & 38.9 & 13.7 & \textit{40.8} & \textbf{45.9} & 17.2 & 20.0  & 18.8 & 14.3 & \textit{26.7} & \textit{34.2} & 55.6 & \textit{35.5} & 37.6 & 30.7 & 19.4 & 16.8 \\
    CTF-Occ \cite{tian2023occ3d} & 3D & \textit{28.53} &\textbf{ 8.1} & \textit{39.3} & 20.6 & 38.3 & 42.2 & 16.9 & 24.5 & \textit{22.7} & 21.0 & 23.0 & 31.1 & 53.3 & 33.8 & 38.  & 33.2 & 20.8 & 18.0 \\
    \noalign{\smallskip}
    \hline
    \noalign{\smallskip}
    SelfOcc \cite{huang2023selfocc} & 2D & 9.30 & 0.0 & 0.2 & 0.7 & 5.5 & 12.5 & 0.0 & 0.8 & 2.1 & 0.0 & 0.0 & 8.3 & 55.5 & 0.0 & 26.3 & 26.6 & 14.2 & 5.6 \\
    OccNeRF \cite{zhang2023occnerf} & 2D & 10.81 & - & 0.8 &  0.8 &  5.1 & 12.5 &  3.5 &  0.2 &  3.1 &  1.8 &  0.5 &  3.9 & 52.6 & - & 20.8 & 24.8 & 18.4 & 13.2 \\
    RenderOcc \cite{pan2023renderocc} & 2D  & 23.93 & 5.7 & 27.6 & 14.4 & 19.9 & 20.6 & 12.0 & 12.4 & 12.1 & 14.3 & 20.8 & 18.9 & \textbf{68.8} & 33.4 & 42.0 & 43.9 & 17.4 & 22.6 \\
    \noalign{\smallskip}
    \hline
    \noalign{\smallskip}
    Ours Base & 2D & 26.14 & 3.2 & 28.8 & \textit{22.2} & 28.0 & 21.7 & 17.2 & 19.6 & 11.0 & 18.0 & 24.1 & 22.0 & \textit{67.3} & 28.7 & 40.0 & 41.0 & 26.2 & 25.6 \\
    Ours Flow & 2D & 28.42 &  1.6 & 27.5 & \textbf{26.0} & 34.0 & 32.0 & 20.4 & \textit{25.9} & 18.6 & 20.2 & 26.0 & 28.7 & 62.0 & 27.2 & 37.8 & 39.5 & \textit{29.0} & \textit{26.8} \\
    Ours Flow & 2D+3D & \textbf{33.86} & \textit{8.0} & 37.6 & \textbf{26.0} & \textbf{42.1} & \textit{42.5} & \textbf{21.6} & \textbf{29.2} & 22.3 & \textbf{25.7} & \textbf{29.7} & \textbf{34.4} & 64.9 & \textbf{37.2} &\textit{44.3} & \textit{43.2} & \textbf{34.3} &\textbf{ 32.5} \\
    \noalign{\smallskip}
    \hline
    \end{tabular}
    \addtolength{\tabcolsep}{2pt}
    }
\end{center}
\end{table*}

Our \textit{Base} model achieves an mIoU score of 26.14\%, outperforming 4 out of 7 Baseline models, although these are supervised by 3D voxel-based labels.
Only FB-Occ~\cite{li2023fb}, TPVFormer~\cite{huang2023tri} and CTF-Occ~\cite{tian2023occ3d} are slightly better.
This shows that volume rendering supervision is sufficient to effectively train an occupancy network, eliminating the need for expensive 3D labels.
Further, our proposed approach exceeds all prior works that also rely on 2D labels.
Specifically, we surpass RenderOcc (23.93\% mIoU) by an increase of almost 10\%, which uses the same backbone and 2D-to-3D transformation.
A more thorough comparison to RenderOcc can be found in \cref{sec:diff_renderocc}.

When adding occupancy flow, the performance of our model is further increased to 28.42\% mIoU, now surpassing all competitors except CTF-Occ, which has a slightly better overall performance.
Compared to RenderOcc we achieve a performance increase by almost 20\%.
The positive effect of occupancy flow is especially observable in the dynamic object classes, as shown in \cref{table:ablation_dynamic}, which compares performance on static and dynamic classes of RenderOcc to our approach.
Methods that use only 2D rendering supervision suffer severely from moving objects, as rendering temporal frames introduces temporal errors.
Already with the \textit{Base} setting, we can increase the mIoU score of the dynamic classes from 16.39\% of RenderOcc to 20.71\%.
We hypothesize that this is due to better supervision of dynamic objects by both filtering wrong rays and disocclusion effects as well as the loss weighting.
Using occupancy flow provides further supervision improvements for dynamic objects since filtering these classes is no longer required.
As expected, this leads to an even greater increase of performance on dynamic objects (mIoU score of 26.44\%).
This experiment shows that, if done correctly, rendering supervision can be effectively used to learn models for 3D semantic occupancy prediction even when dynamic objects are present.

\begin{table}
\small
\begin{center}
 \caption{
            \textbf{Comparing mIoU between dynamic and static classes.}
            Using flow, predictions for the dynamic classes are improved significantly (best performance in \textbf{bold}).
        }
        \label{table:ablation_dynamic}
        \begin{tabular}{l@{\hspace{.5em}}c@{\hspace{.5em}}c@{\hspace{.5em}}c@{\hspace{.5em}}}
            \hline
            \noalign{\smallskip}
            \multirow{2}{*}{Method} & \multirow{2}{*}{mIoU} &  mIoU & mIoU \\ & & Static & Dynamic\\
            \noalign{\smallskip}
            \hline
            \noalign{\smallskip}
            RenderOcc \cite{pan2023renderocc} & 23.93 & 30.63 & 16.39\\
            \noalign{\smallskip}
            \hline
            \noalign{\smallskip}
            Ours Base & 26.14 (+2.21) & \textbf{30.97} (+0.34) & 20.71 (+4.32) \\
            \noalign{\smallskip}
            Ours Flow & \textbf{28.42} (+4.49) & 30.18 (-0.45) & \textbf{26.44} (+10.05)\\
            \hline
            \noalign{\smallskip}
        \end{tabular}
    \end{center}
\end{table}

Finally, we provide results of our model when it is also trained with 3D labels, combining it with rendering supervision.
Adding 3D supervision increases the performance yielding up to 33.86\% mIoU, outperforming all prior methods significantly, thus setting the new state-of-the-art performance for the Occ3D-nuScenes dataset.

\begin{figure*}
  \centering
      \includegraphics[page=2, trim=0cm 3.6cm 4.16cm 0cm, clip, width=1\textwidth]{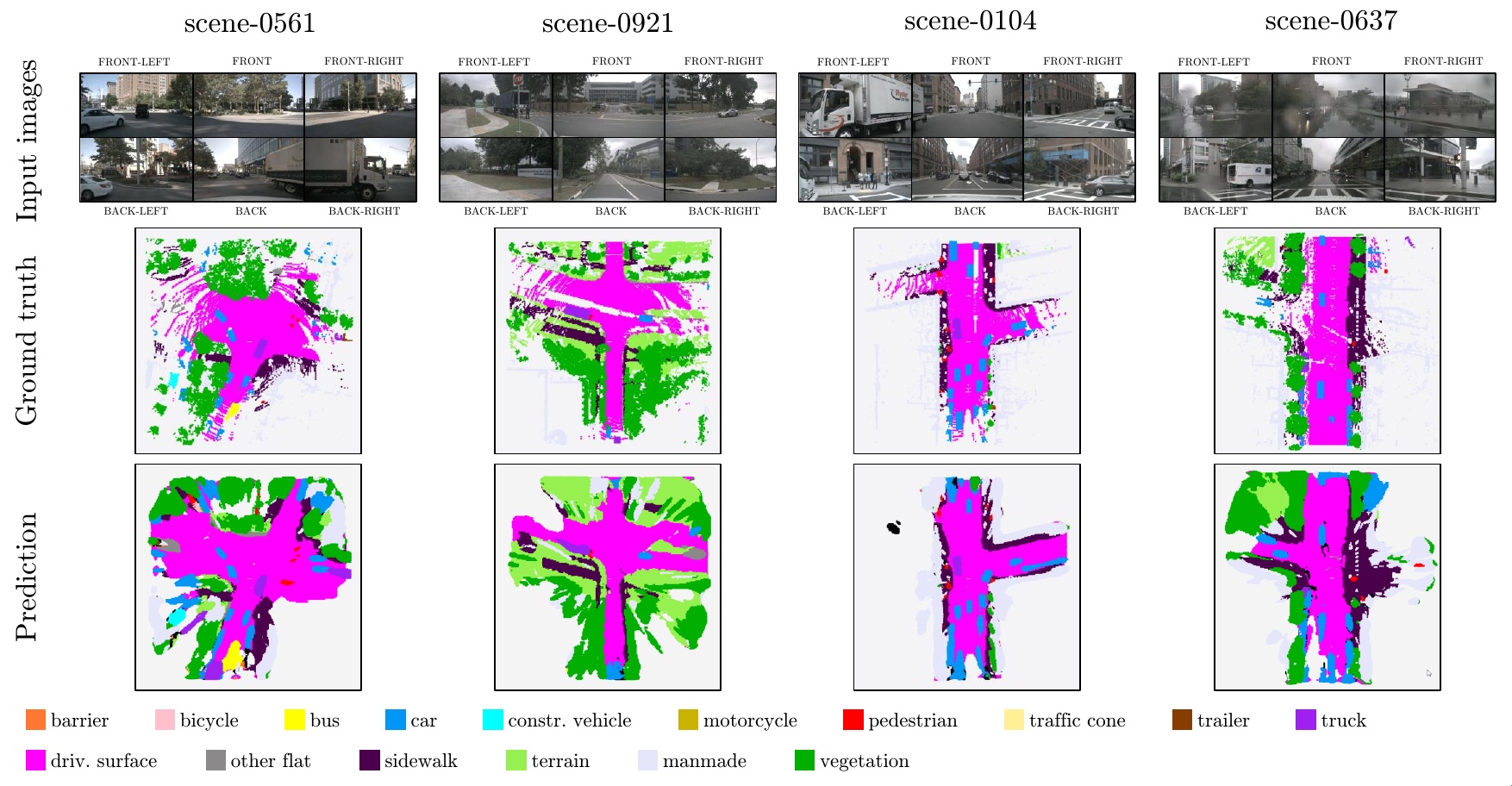}
  \caption{
  \textbf{Qualitative results on the Occ3D-nuScenes dataset, viewed from the top.}
  The proposed model can estimate static and dynamic objects in 3D correctly and generalizes well to unseen areas.
The shown results are generated using the \textit{Ours Flow 2D} model.}
   \label{fig:qualitative}
\end{figure*}

Qualitative results on the Occ3D-nuScenes dataset are shown in \cref{fig:qualitative}, highlighting the ability of the proposed approach to estimate 3D scene geometry and dynamic objects correctly.
Further qualitative results can be found in \cref{app:qual}.

\subsection{Ablation Study}\label{sec:exp3}

We provide further insides into the different components of OccFlowNet in order to show their influence on the performance.
All ablation experiments are conducted with the \textit{Base} setting, e.g., without using occupancy flow.

\paragraph{Proposed Components}
\Cref{table:ablation2d} shows the effect of the different components of the proposed model on the overall performance.
We train only with 2D labels, and add components one-by-one.
When using semantic supervision only, 15.19\% mIoU can be achieved, however, adding depth supervision already improves the performance to 21.79\%.
Adding temporal rendering increases the mIoU by another $+3\%$, however, dynamic objects are not accounted for yet.
When we filter out dynamic objects and add the loss weighting to account for less supervision of dynamic objects, an mIoU score of 26.14\% is achieved.

\begin{table}
\small
\centering
\caption{Ablation study investigating different settings of the proposed model. 
All methods are trained only with 2D labels. 
\textit{LW} stands for "Loss weighting".}
\setlength{\tabcolsep}{0.015\linewidth}
\begin{tabular}{ccccc}
\hline
\noalign{\smallskip}
\multirow{2}{*}{Semantics} & Semantics & Temporal & Dynamic filter & \multirow{2}{*}{mIoU} \\
& + Depth & Rendering & + LW \\
\noalign{\smallskip}
\hline
x & & & & 15.19\\
x & x & & & 21.79\\
x & x & x & & 24.97\\
x & x & x & x & 26.14\\
\hline
\end{tabular}
\label{table:ablation2d}
\end{table}

\paragraph{Time Horizon}
We ablate the time horizon $O$ in the range of $\{0, 1, ..., 5\}$ to show how temporal rendering leads to better overall supervision. 
Note that the horizon only influences the temporal rendering during training.
During inference, rendering is not performed.
The set of available frames to the model is always $I = \{ I_{-O}, ..., I_{O} \}$.
As can be seen in \cref{table:ablation_horizon}, when $O=0$, no temporal rendering is used and a performance of 21.79\% mIoU is achieved.
Adding a single past and future frame already improves the performance to 24.70\% mIoU (outperforming RenderOcc at $O=3$).
Increasing the horizon further improves the performance up to $O=3$.
Afterwards, the performance declines again.
This is likely due to the fact that the cameras are now too far away from each other and rays start to intersect less again, eventually reducing the effective supervision.

\begin{table}
\small
\centering
\caption{Performance of OccFlowNet using different time horizons $O$. 
The best performance is achieved with $O=3$, shown in \textbf{bold}.}
\label{table:ablation_horizon}
\setlength{\tabcolsep}{0.015\linewidth}
\begin{tabular}{c|cccccc}
\hline
\noalign{\smallskip}
Time Horizon & 0 & 1 & 2 & \textbf{3} & 4 & 5\\
\noalign{\smallskip}
\hline
\noalign{\smallskip}
mIoU & 21.79 & 24.70 & 25.19 & \textbf{26.14} & 25.89 & 25.55 \\
\noalign{\smallskip}
\hline
\end{tabular}
\end{table}

\paragraph{Comparison to RenderOcc} \label{sec:diff_renderocc}
To further explore the performance of our model compared to RenderOcc, we identify key differences between these approaches and train our model with settings identical to RenderOcc.

The first difference lies in the choice of loss function.
The authors of RenderOcc propose to use the SILogLoss~\cite{eigen2014depth} as the $\mathcal{L}_{depth}$, which is suitable when metric depth is unavailable.
However, as LiDAR points projected on images provide depth on a metric scale, we can use the mean squared error loss to learn metrically accurate depth.
Next, we estimate the density $\hat{V}_\sigma$ as probabilities using a sigmoid function, whereas RenderOcc does not bound the density from above.
This does not allow to interpret $\hat{V}_\sigma$ as probabilities, which makes thresholding harder and likely leads to a different optimization process.
Finally, the treatment of dynamic objects differs between the models. 
RenderOcc ignores dynamic rays in temporal time steps, but do not handle disocclusions.
In addition, no loss weighting is used but instead, the authors propose the \textit{weighted ray sampling}, which weights the probability of rays being sampled based on the class occurrence.
To enable a direct comparison to RenderOcc we adapted our implementation using the settings of RenderOcc.
The results of this comparison are given in \cref{table:ablation_renderocc}.
It is observable that we can roughly reproduce the results of RenderOcc using their setting in our implementation.
We conclude that our setting is superior compared to RenderOcc.

\begin{table}
\begin{center}
\caption{Performance of OccFlowNet Base vs when using the proposed setting of RenderOcc.
Our setting clearly outperforms the competitors setting, demonstrating its potency.}
\label{table:ablation_renderocc}
			\begin{tabular}{l@{\hspace{4em}}l@{\hspace{1em}}}
				\hline
                \noalign{\smallskip}
				Method & mIoU \\
				\noalign{\smallskip}
				\hline
				\noalign{\smallskip}
				Ours Base & 26.14\\
				\noalign{\smallskip}
				Ours RenderOcc setting & 23.31 (-2.83) \\
				\noalign{\smallskip}
				\hline
			\end{tabular}
\end{center}
\end{table}
\section{Conclusion and Future Work}\label{sec:conclusion}
In this work, we demonstrate the feasibility of training a 3D occupancy model using volume rendering, leveraging only 2D labels.
By incorporating temporal rendering and dynamic object handling via occupancy flow, OccFlowNet is the first 2D supervised approach to achieve competitive performance compared to 3D-based models, while surpassing concurrent 2D-based methods.
Therefore, OccFlowNet closes the gap between 3D and 2D supervised methods,
enabling training without expensive 3D voxel labels, marking a significant advancement towards self-supervised, purely vision-based methods.
Combining both 2D and 3D supervision further improves the performance of OccFlowNet outperforming all previous occupancy estimation models.

In future research, the exploration of self-supervised learning can be extended by investigating vision-based methods for acquiring 2D labels, such as leveraging strong foundation models for depth and semantic labels, thereby eliminating the reliance on LiDAR sensors.
Furthermore, additionally estimating occupancy flow with the model and jointly optimizing it with volume rendering can be investigated, to become independent of labeled 3D boxes, which is a major drawback of the proposed model.
Moreover, the estimation of color features within the voxel grid and the utilization of photometric losses for model training, akin to traditional NeRF approaches~\cite{mildenhall2021nerf}, will be a promising direction.

{
    \small
    \bibliographystyle{ieeenat_fullname}
    \bibliography{main}
}

\appendix
\renewcommand\thefigure{\thesection.\arabic{figure}}
\setcounter{figure}{0}
\section{Qualitative Results}\label{app:qual}

\subsection{Rendered Depth and Semantics}
\begin{figure*}
  \centering
      \includegraphics[page=3, trim=0cm 0.86cm 3.08cm 0cm, clip, width=1\textwidth]{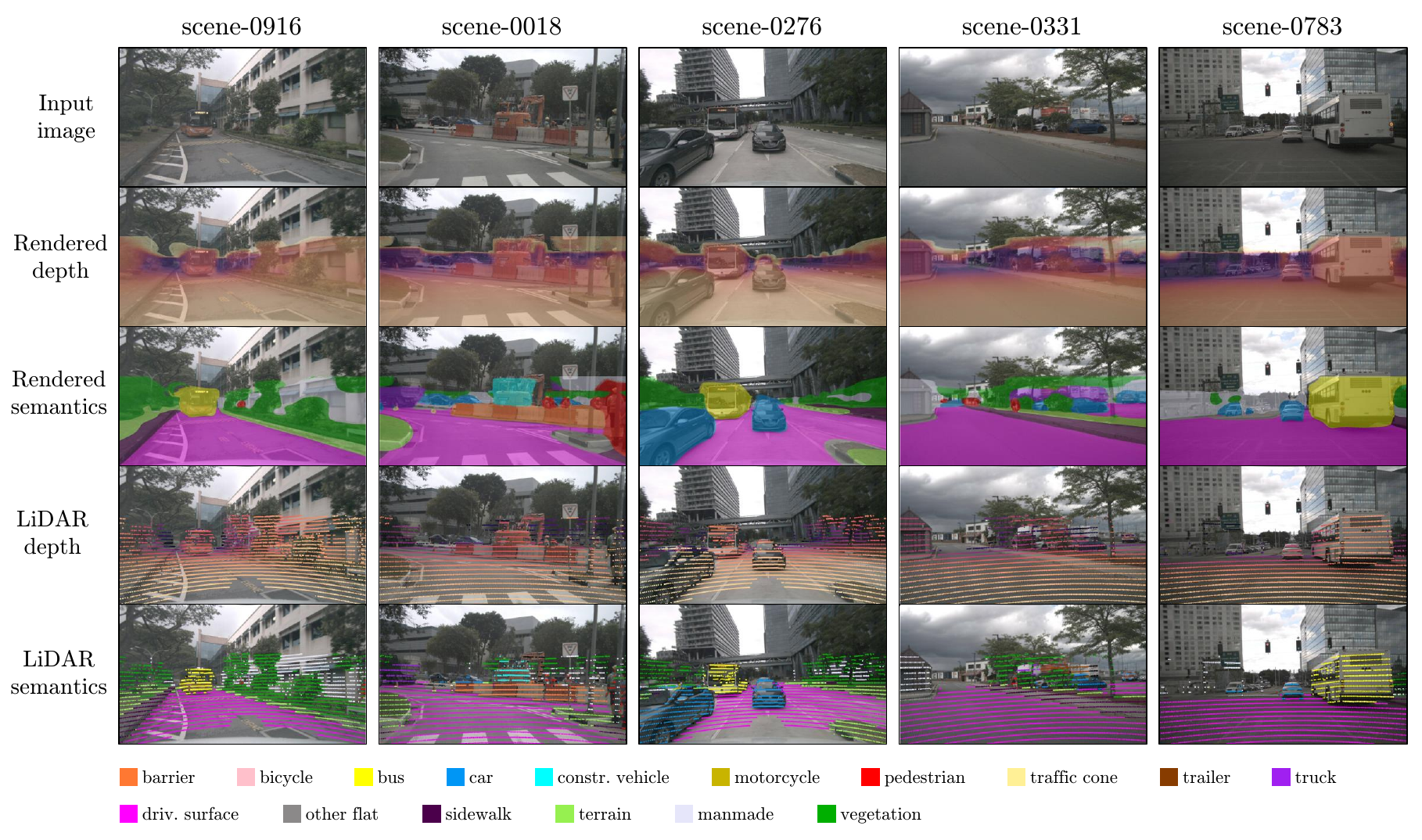}
  \caption{
  \textbf{Qualitative results on the Occ3D-nuScenes validation set.}
  Each column shows an \textit{input image}, and the corresponding \textit{rendered depth} $\hat{D}$ and \textit{rendered semantics} $\hat{S}$ when rendering the occupancy predictions using volume rendering, simulating the training process.
  Below, the 2D training labels are shown, generated by projecting annotated LiDAR point scans onto the input images.
  }
\label{fig:qual_render}
\end{figure*}

Additional qualitative results are given in \cref{fig:qual_render}.
The figure shows what the model "sees" during training, when the 3D voxel predictions are rendered into the 2D space using differentiable volume rendering.
As clearly visible, the model has learned to estimate depth and semantics correctly, and that volume rendering can be effectively used to create 2D predictions, while being fully differentiable.

\subsection{Predicted Occupancy}
\begin{figure*}
  \centering
      \includegraphics[page=6, trim=0cm 2.84cm 3.87cm 0cm, clip, width=1\textwidth]{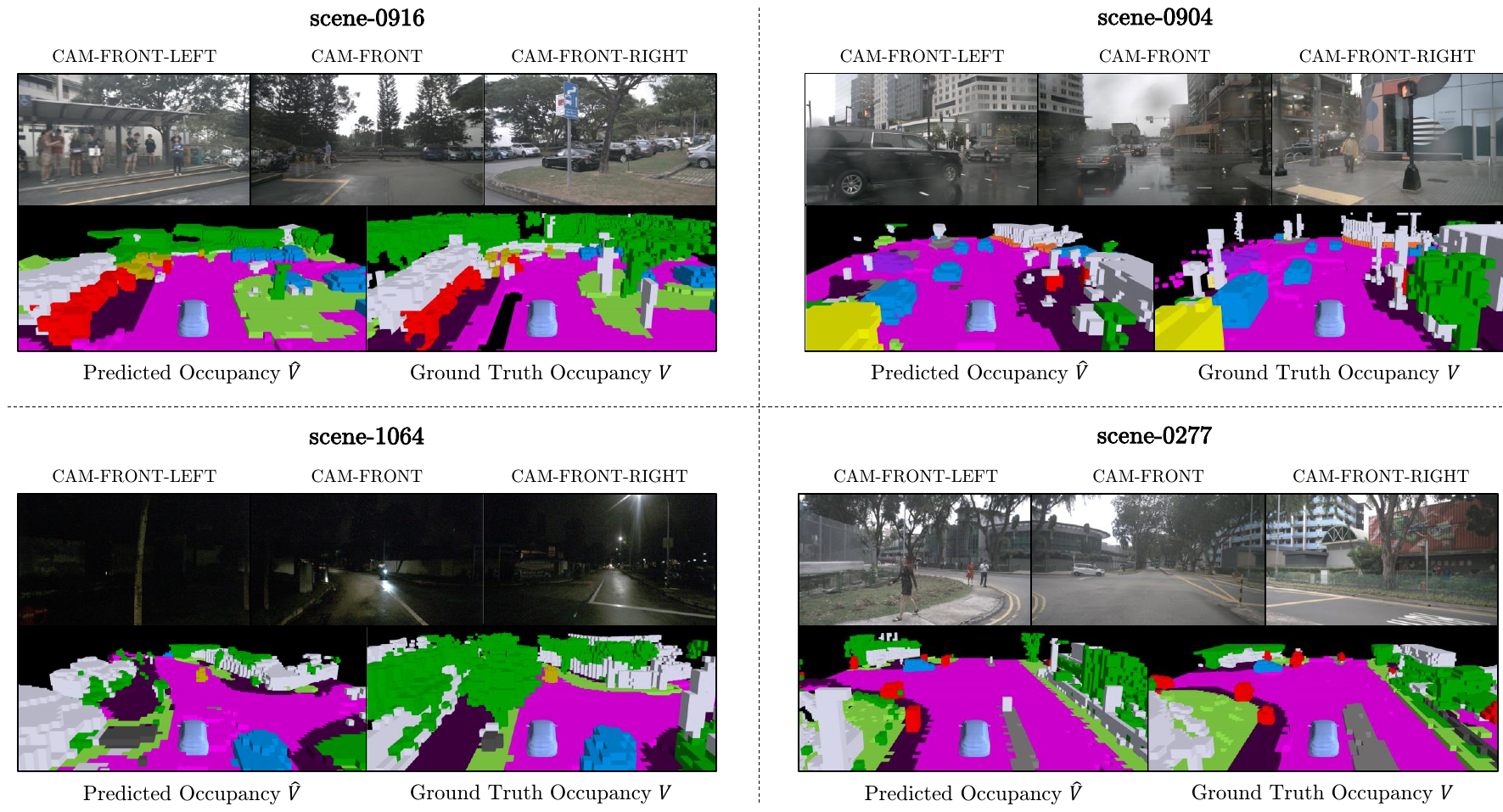}
  \caption{
  \textbf{Qualitative results on the Occ3D-nuScenes validation set.}
  Images of the occupancy space are taken from above and behind the ego vehicle ("third-person" view).
  The model can estimate the semantic occupancy well compared to the ground truth.
  }
\label{fig:qual_voxel}
\end{figure*}

In \cref{fig:qual_voxel}, we show examples of predicted occupancy of our OccFlowNet model compared to ground truth occupancy.
The figure depicts results on the Occ3D-nuScenes validation set, and the model used is OccFlowNet with occupancy flow.
As observable, the model can accurately estimate dynamic and static objects around the vehicle, in day scenes as well as in night scenes.

\section{Details: Dynamic Ray Filter}
\begin{figure*}
  \centering
      \includegraphics[page=4, trim=0cm 8.31cm 15.23cm 0cm, clip, width=0.9\textwidth]{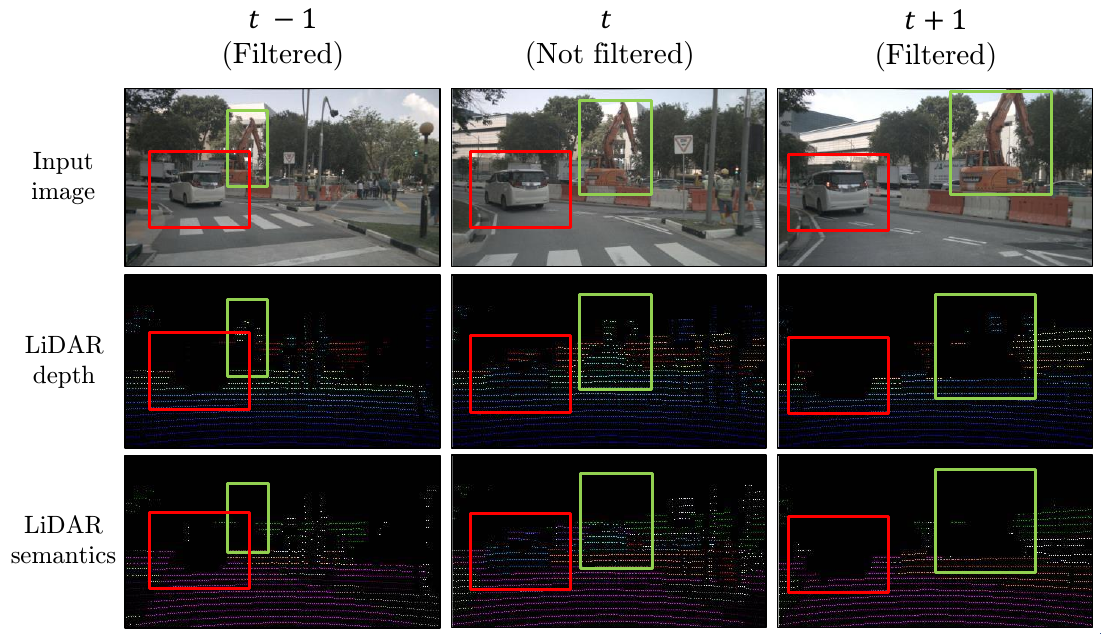}
  \caption{
  \textbf{Illustration of the dynamic ray filter.}
  Shown are the \textit{input image}, \textit{LiDAR depth} and \textit{LiDAR semantics} for a single frame across three consecutive time steps when using the dynamic ray filtering (\cref{sec:ray_filter}).
  LiDAR points corresponding to dynamic objects are removed in adjacent time steps $t-1$ and $t+1$.
  In the current time step $t$, points of dynamic objects are preserved.
  }
\label{fig:dynamic_ray_filter}
\end{figure*}

\Cref{fig:dynamic_ray_filter} demonstrates how the dynamic ray filter introduced in \cref{sec:ray_filter} is applied.
During training, while loading the data, we filter out dynamic objects in adjacent time steps.
LiDAR points associated with dynamic object classes are removed, such as the \textit{car} within the red box or the \textit{construction vehicle} within the green box depicted in the figure.
In the current time step, dyanmic LiDAR points are kept, as the dynamic objects are at the correct position in the current frame.
Then, a ray is generated for each remaining LiDAR point on each input image as described in \cref{sec:ray_gen}.
It is important to note that, as explained in \cref{sec:disocclusions}, this technique alone is insufficient, as moving objects may expose background points that are not filtered, introducing misleading supervision.

\section{Details: Occupancy Flow}
\begin{figure*}
  \centering
      \includegraphics[page=5, trim=0cm 9.84cm 5.58cm 0cm, clip, width=1\textwidth]{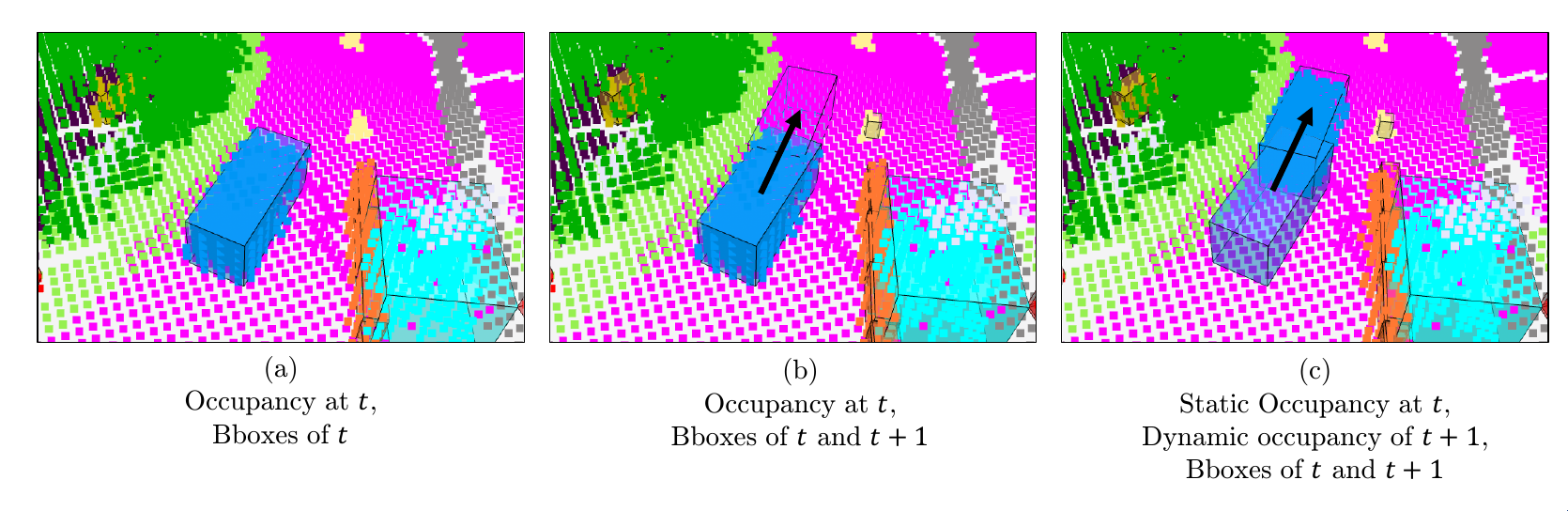}
  \caption{
  \textbf{Example of the occupancy flow.}
  We precompute the transformations between corresponding boxes in adjacent time steps, and use them to relocate the estimated occupancy $\hat{V}$ to the target time steps.
  }
\label{fig:occ_flow}
\end{figure*}

\Cref{fig:occ_flow} illustrates how the occupancy flow (\cref{sec:occ_flow}) is applied during training, using the example of a single instance of the \textit{car} class.
Given the current time step $t$ and a target time step $t+1$, we utilize the precomputed transformation from the current bounding box to the target bounding box to move the estimated occupancy to the target time step.
This procedure is repeated for all adjacent time steps within the specified horizon.
Consequently, the dynamic ray filter becomes unnecessary, as dynamic objects are now correctly positioned even in temporal time steps.
Significantly more rays can be generated for dynamic objects, greatly increasing their detection performance.

\end{document}